# Arbitrary Style Transfer with Structure Enhancement by Combining the Global and Local Loss

Lizhen Long[1] and Chi-Man Pun[1*]

[1]Department of Computer and Information Science, University of Macau, Macau, China.

*Corresponding author(s). E-mail(s): cmpun@umac.mo;
Contributing authors: mb95426@um.edu.mo;

**Abstract**

Arbitrary style transfer generates an artistic image which combines the structure of a content image and the artistic style of the artwork by using only one trained network. The image representation used in this method contains content structure representation and the style patterns representation, which is usually the features representation of high-level in the pre-trained classification networks. However, the traditional classification networks were designed for classification which usually focus on high-level features and ignore other features. As the result, the stylized images distribute style elements evenly throughout the image and make the overall image structure unrecognizable. To solve this problem, we introduce a novel arbitrary style transfer method with structure enhancement by combining the global and local loss. The local structure details are represented by Lapstyle and the global structure is controlled by the image depth. Experimental results demonstrate that our method can generate higher-quality images with impressive visual effects on several common datasets, comparing with other state-of-the-art methods.

# 1 Introduction

Broadly speaking, artistic style transfer is a creative technique used to generate an artistic image which combines the structure of a content image and the artistic style of the artwork. Nowadays, many impressive results have been achieved by different way. But it should be noted that image structure representations are important elements in producing impressive visual image.

Traditionally, style transfer has been studied as a texture synthesis task. Based on the classification of previous research [1] on style transfer, it can be roughly divided into two types: Image-Optimization methods and Model-Optimization methods. Image-optimization methods iteratively optimize stylized image by various network. Gatys et al.[2] was the first to apply Convolutional Neural Networks to the style transfer task, which shows that the pre-trained classification network can extracted the correlations between features to capture the style patterns well and iteratively generates stylized images. The Gatys et al.[2] method is flexible enough to combine the content and the style of the given images, but it also has its limitations that the optimization process is prohibitively slow. Following works improve of [2], Li et al. [3] theoretically proved that the Gram matrix matching procedure is equivalent to minimizing MMD using a second-order polynomial kernel in NST. E. Risser et al. [4] used histogram losses to synthesize textures and Nicholas et al. [5] proposed style transfer by Relaxed Optimal Transport and Self-Similarity (STROTSS), which can both achieved a particular visual effect. However, the problem of Image-optimization methods restricted by the slow online optimization process are fundamentally not well-solved. At the same times, Model-Optimization methods reconstruct stylized image through the feed-forward network which mainly subdivided three types. (1) Per-Style-Per-Model methods [6][7][8][9] are trained a single transfer network for each desired style. (2) Multi-Style-Per-Model methods [10][11] [12][13] are trained a single transfer network to produce images in multiple styles and even blend more than one style together. (3) Arbitrary-Style-Per-Model methods [14][15][16][17][18] can take a content image and a style image as input and perform style transfer in a single, feed-forward pass. In other words, it can extract and apply any style to an image.

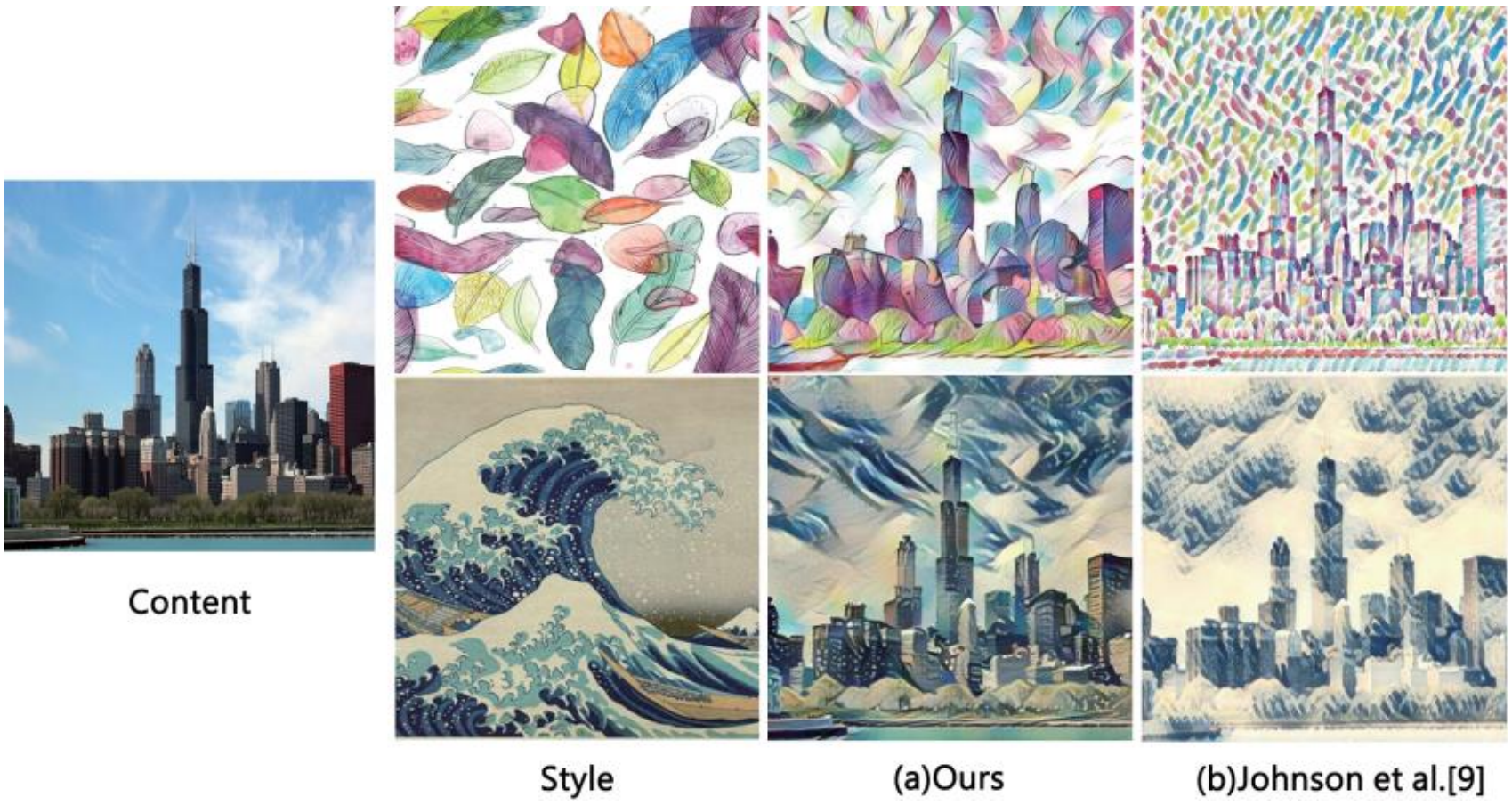

**Fig. 1** The distributed style textures destroy the local detail and the original structure in content images by Johnson et al.'s method [6] as in column (b). Our method preserves the structure of the original image and the spatial distribution as in column (a), and the visual results are better than [6]. Our method is arbitrary style transfer, while Johnson method is per-style-per style transfer.

Reviewing these methods, we find that some of these methods ignore the inherent structural information of the image. The results are not satisfactory that the stylized images distribute style elements evenly throughout the image when the content image with complex spatial information, which makes the overall structure of the image unrecognizable. Seen the example in Fig.1, the distributed textures destroy the local detail and the original structure. These problems are caused by the poor transfer of the structural representation of the image. According to this observation, we can find that the features of the pre-trained classification network middle layers affect the image structure representation. However, the traditional classification networks were designed for classification which usually focus on high-level features and ignore other features. To solve the above-mentioned limitations, we propose an improved arbitrary style transfer method with structure enhance, which can generate high-quality stylized images and preserving more content structure details. This is achieved by a style-attentional network (SANet) and a new loss function combination. The Loss functions contain the Laplacian loss function and the depth loss function. The depth map [19] can well reflect the global spatial structure of image. We compute the depth loss through the depth map, which preserves the overall image layout during style transfer. On the other hand, the Laplacian loss can preserve the local detail structures of the content image and reduce the generation of artifacts, which has been proved by [20]. Our experiment results show that this method can generate high quality image that preserves the structure and details of the original images.

The main contributions of our work can be summarized as follows:

1. We introduce a new structure-enhancement image framework for arbitrary style transfer, which improves the stylized image quality under the guidance of a combination of global structure loss and a local structure loss.
2. We show that image structure representation plays a very important role in style transfer, and more visually striking stylized results can be produced by designing a suitable image structure representation.
3. Our experiments demonstrate that our method can generate (at about 17–24 frames per second for 512 x 512 pixels) stylization results with highly efficient and high quality, where global and local structure patterns are both effectively synthesized.

# 2 Related work

**Arbitrary Style Transfer.** The main goal of Arbitrary Style Transfer is to transfer images of any given style by using only one trained network, with the highest possible efficiency and generation quality. Many existing studies[3][8][10][15][16][18][14][21][22] have implemented improvements in the trade-off between quality and efficiency in Style Transfer. Li et al.[3]proved that the batch normalization (BN) statistic representation can be used the aligning mean and variance of feature maps in a certain layer to represent the style. Ulyanov et al. [8]found that the instance normalization (IN) was applied to feed-forward texture synthesis and stylization networks, which improved the quality and diversity of the generated results. Dumoulin et al.[10] proposed the conditional instance normalization (CIN) approach that set the different affine parameters of the instance normalization for each style to model the different style. However, previous methods still cannot fully implement arbitrary style transfer. To solve the limitations of previous problem, Huang et al. [15] proposed the groundbreaking adaptive instance normalization (AdaIN) algorithm that effectively adjust the mean and variance of the content image to match the mean and variance of the style image by transferring global feature statistics. AdaIN layer is as simple as an IN layer, but its method still lacks in quality. Besides, there are also other improved arbitrary style transfer methods, Li et al [16] embedded a pair of feature transforms (whitening and coloring) to an image reconstruction network that WTC reflect the direct matching of feature covariance of the content image to the given style image. The styleBank method [11] train the filtering kernels to adding the new image style. Lin et al. proposed the Avatar-Net method [18] , which key ingredient is a style decorator that can transfers the content features to the semantically nearest style features and preserve detailed style patterns. Park et al.[14] proposed Style-Attentional Networks that embed local style patterns in content feature map with the aid of style attention mechanism. The neural style transfer method is becoming more and more powerful. However, the quality of the generated image still needs more improvement and Image structure representations improvement is one of the important directions. In this work, we propose a new structure enhanced for image structure representation in arbitrary Style Transfer.

**Laplacian Loss.** In computer vision, Laplacian Matrix is widely used to detect edges and contours. It can be used to produce by a Laplacian operator and reflect the detailed structures of the image. When the Laplacian Matrix applied in style transfer, Li et al.[20] found that the deviation in the Laplacian matrix can correspond to unexpected distortions or artifacts in the stylized image. Therefore, The Laplacian loss is used to measure the difference in Laplacian matrix between the content image and the stylized image to reflect the corresponding difference in detail structure, which called as "Lapstyle". Besides, Lin et al [23] also applied the Lapstyle to Drafting and Revision network for fast artistic style transfer. In our work, the stylized image can preserve the detail structures by the Laplacian Loss [24] in arbitrary style transfer.

**Deep Image Representations.** Torralba et al. [19]pointed out that the depth of the scene can be recognized the structures present in the image. The Depth map can effectively reflect the spatial distribution and the global structure of the image. The previous works [25][26][27][22][28] depth preservation as additional loss to overall image layout in style transfer. Inspired by this work, we use the method of Moodepth model [29][30][31] to calculate the deep loss, which can well reflect the global structure of the image.

# 3 Method

In this section, we proposed style transfer system which is composed of two main parts: an encoder–decoder module and a representation module are used to define four loss functions. As shown in Fig.3, the generator network use SANet [14] to generate high-quality stylized images. The loss functions are designed to enhance the structure of the content & style that appropriately reflect global and local structure of the image.

## 3.1 Encoder–Decoder Module

As shown in Fig.3, the proposed style transfer system takes image $I_c$ and an arbitrary style image $I_s$ as inputs to generate synthesizes a stylized image $I_{cs}$ which combined semantic structures of $I_c$ with style characteristics of $I_s$. In this work, the generator network includes three parts: encoder, Style-Attention Network, a symmetric decoder. We use the pretrained VGG-19 network [32]as encoder and the decoder follows the settings of [15].

To generate high-quality stylized images, we utilize the layers (Relu_4_1 and Relu_5_1) of the encoder E (the pretrained VGG-19) to encode $I_c$ and $I_s$ into the feature space. After extracting their feature maps pair $F_c = E\ (I_c)$ and $F_s = E\ (I_s)$ respectively, there is a parallel SANet designed to refine the feature maps separately. We input both feature maps to the SANet module, which maps the correspondence between the content feature map $F_c$ and the style feature map $F_s$, producing the following output feature maps:

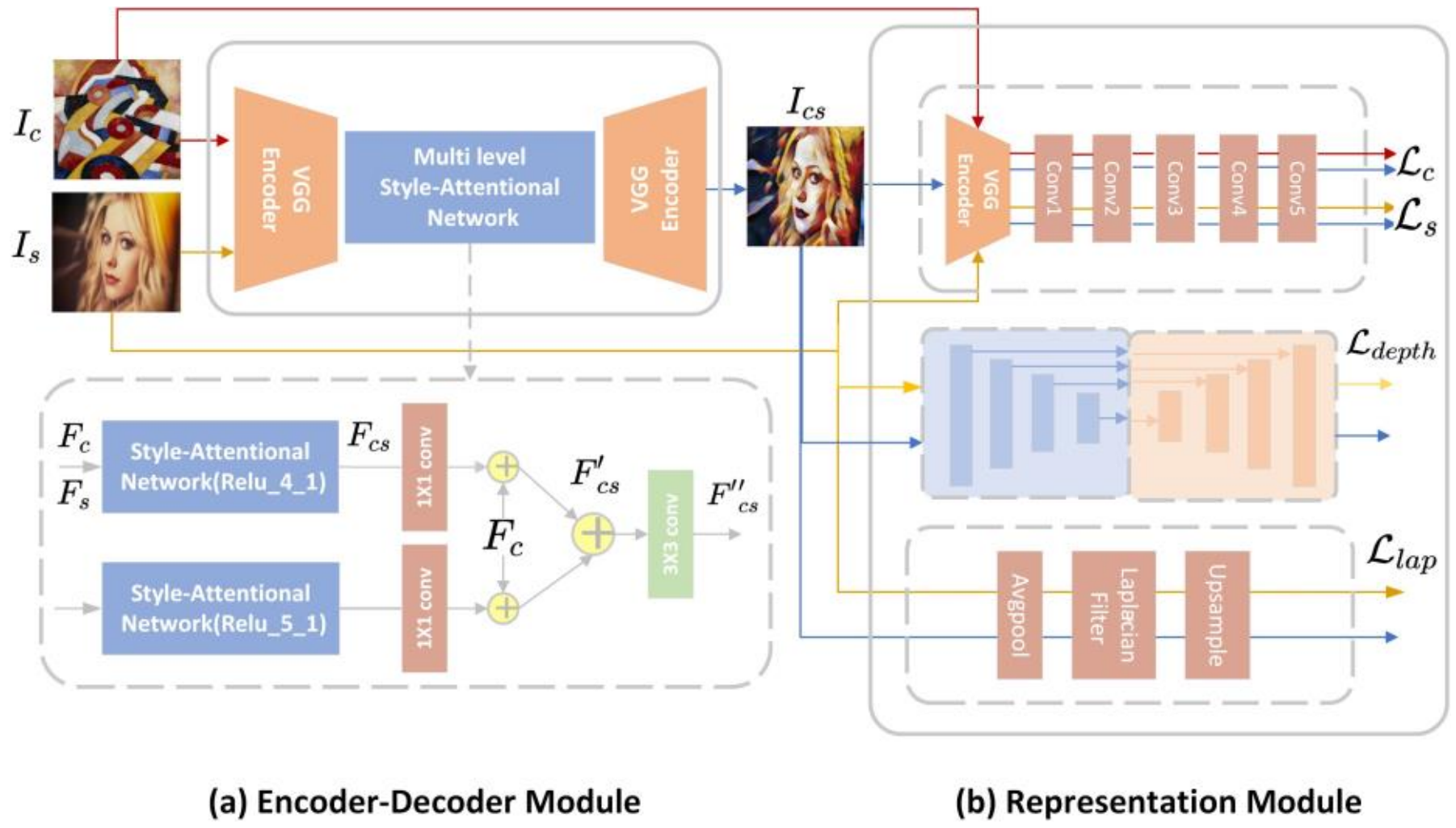

**Fig. 3** The proposed network architecture for Arbitrary Style Transfer with Structure Enhancement by Combining the Global and Local Loss. (a) The Encoder-Decoder Module. (b) The Representation Module.

$$F_{cs} = SANet(F_c, F_s) \tag{1}$$

Summing the following two matrices $F_c$ and after applying $1 \times 1$ convolution $F_{cs}$ to generate $F_{cs}$. the pair of the output feature map from the parallel SANet as:

$$F'_{cs} = F_c + W_{cs}Conv_{1\times1}(F_{cs}) \tag{2}$$

$$F''_{cs} = Conv_{3\times3}(F'^{R_4_1}_{cs} + F'^{R_5_1}_{cs}) \tag{3}$$

where "+" means element-wise summation.$F''^{R_4_1}_{cs}$ and $F''^{R_5_1}_{cs}$ are the output feature maps obtained from the parallel SANets.

Then, the stylized output image $I_{cs}$ is synthesized by sending the $F''cs$ into the decoder D as follows:

$$I_{cs} = D(F''_{cs}) \tag{4}$$

## 3.2 Representation module

As shown in Fig.3, The representation module includes four loss functions: $L_c$,$L_s$, $L_{lap}$ and $L_{depth}$, where $L_c$ and $L_s$ correspond to the style loss and content loss. $L_{lap}$ and $L_{depth}$ are based on the Structure-Representation and correspond to the depth loss and edge loss. $L_{lap}$ and $L_{depth}$ represent the loss function of the global structure and the local structure, respectively. The total loss is defined as:

$$L_{total} = \lambda_c L_c + \lambda_s L_s + \lambda_{lap} L_{lap} + \lambda_{depth} L_{depth} \tag{5}$$

where $\lambda_c$, $\lambda_s$ , $\lambda_{lap}$ and $\lambda_{depth}$ are the weights of different corresponding losses.

### 3.2.1 Representation Loss

Following the approach in [2], we use the pretrained VGG-19 to compute two loss functions $L_{content}$ and $L_{style}$ for training decoder. Specifically, $L_{content}$ is defined as the Euclidean distance (squared, normalized) between the output feature $F_c$ and the features of the stylized output image $E\ (I_{cs})$ in selected layers (Relu_4_1 and Relu_5_1) as:

$$L_c = |E\ (I_{cs}) - F_c|_2 \tag{6}$$

Similar to [15], The style loss is defined as follows:

$$L_s = \sum_{i=1}^{L} ||\mu(\phi_i(I_{cs})) - \mu(\phi_i(I_s))||_2 + ||\sigma(\phi_i(I_{cs})) - \sigma(\phi_i(I_s))||_2 \tag{7}$$

where each $\phi_i$ represents the feature of the layer in the encoder used to compute the style loss. In the experiments, we use Relu_2_1, Relu_3_1 , Relu_4_1 and Relu_5_1 layers with equal weights.

### 3.2.2 Structure-Representation Loss

The structure representation loss consists of two loss: a global structure loss $L_{depth}$ and a local structure loss $L_{Lap}$. Their advantages compensate for the deficiencies of content & style loss in capturing and maintaining structure.

Inspired by[20], Lapstyle are particularly appropriate to represent the local structure. the Laplacian filter is defined as:

$$L = \begin{bmatrix} 0 & -1 & 0 \\ -1 & 4 & -1 \\ 0 & -1 & 0 \end{bmatrix} \tag{8}$$

The Laplacian matrix ("Laplacian" in short) of an image is obtained by convolving the input image with Laplacian filter. Since the input image has three channels, the final Laplacian should be the sum of the absolute values of the three-channel Laplacian. As for the Laplacian loss, we compute the MSE loss between the Laplacian matrices of the stylized image $I_{cs}$ and the content image $I_c$ respectively in three channels:

$$\mathcal{L}_{lap} = \sum_{ij} \left(L\left(I_{cs}\right) - L\left(I_c\right)\right)^2_{ij} \tag{9}$$

$$\begin{aligned} \mathcal{L}_{lap} = & \sum_{ij} \left(L\left(I^R_{cs}\right) - L\left(I^R_c\right)\right)^2_{ij} + \sum_{ij} \left(L\left(I^G_{cs}\right) - L\left(I^G_c\right)\right)^2_{ij} \\ & + \sum_{ij} \left(L\left(I^B_{cs}\right) - L\left(I^B_c\right)\right)^2_{ij} \end{aligned} \tag{10}$$

where i and j represent the pixels of the image.

The depth map is an important feature of the image and is very suitable to reflect the global structure, since it contains 3D feature information and location information about the object. The global structure extraction of the image adopts the general model of self-supervised monocular depth estimation [29]. The entire image is used as input and the image depth is directly predicted. $\mathcal{L}_{depth}$, which stands for the global-structure difference of two images $I_{cs}$ and $I_c$, is calculated the Euclidean distance between the image $I_{cs}$ and the content image $I_c$ in the same way as content loss:

$$\mathcal{L}_{depth} = \left|Depth\left(I_{cs}\right) - Depth\left(I_c\right)\right|_2 \tag{11}$$

where $Depth\left(I_{cs}\right)$ and $Depth\left(I_c\right)$ represent the depth map of $I_{cs}$ and the depth map of $I_c$.

# 4 Experimental Results

As shown on Fig.3, the overview of our network based on the proposed network architecture for arbitrary style transfer with structure enhancement by combining the global and local Loss. In this section,we will show the experiments and results.

## 4.1 Settings

We trained our proposed arbitrary style transfer network by using MS-COCO [33] as the content images dataset and WikiArt [34] as the style images datasets, which both datasets contain roughly 80000 training images. We used the Adam optimizer [35] 0.0001 as self-adjusted learning rate and a batch size of 5 content-style image pairs. During training, we first resize the smaller dimension of both images to 512 while preserving the aspect ratio, and then randomly crop the 256 × 256 pixels region size. During the testing, our network can input any image size, because the trained network is a fully convolution.

## 4.2 Comparisons

In this section, we compare our method with SANet [14] proposed by Park et al and other three types of arbitrary style transfer method: 1) the flexible but slow iterative optimization method proposed by Gatys et al.[2]. 2) the different kind of feature transformation-based methods (such as WCT [16] proposed by Li et al. and AdaIN [15] proposed by Huang et al. 3) the patch-based method Avatar-Net proposed by Sheng et al.[18] .If not mentioned otherwise, some results are taken directly from their paper to be fair. All the test images are of size 512X512.

### 4.2.1 Qualitative Example

As shown in Fig.5, we show the example experimental results of state-of-the-art arbitrary style transfer method, which compared with our method. Note that none of the test style images and content images were trained on the network. We also provide the additional test results in Fig.4

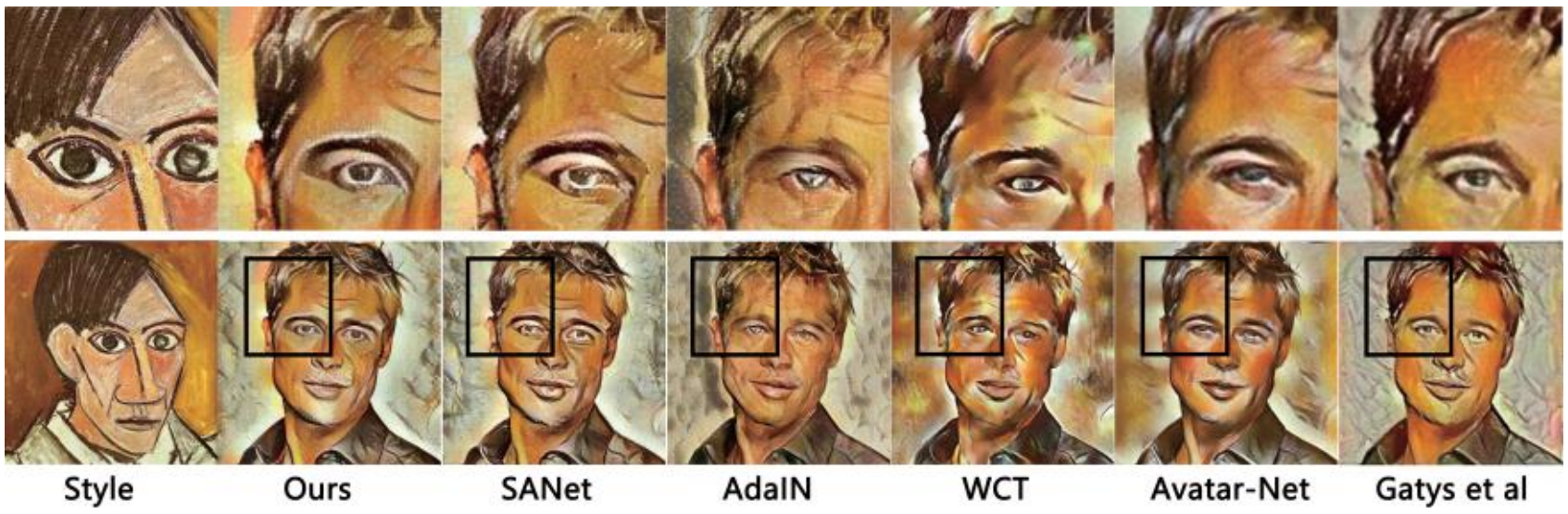

**Fig. 4** Comparisons on visual quality with other style transfer methods, our results can preserve the more detail specially on the human eyes and hair texture details. Regions marked by bounding boxes in the bottom row are enlarged in the top row for better visualization.

The Gatys et.al.[2] method is based on iterative optimization to achieve arbitrary style transfer. However, it sometimes encounters bad local minima in the process of iteration resulting in not so well styled image (e.g., row 3 and 4 in Fig.5). Due to AdaIn[15] synthesizes stylized images by matching the mean and variance between adjusted content and style features. Its result sometimes preserves some of the color distribution of the content image (e.g., row 1 and row 8 in Fig.5) and phenomena that cause the local style distortions and artifacts (e.g., row 2 and row 3 in Fig.5), which can affect the visual of the resulting image. We can see our method result compared with the AdaIN method, our proposed method is able to eliminate unexpected artifacts (e.g., row 3 and row 4 in Fig.4). In addition, WCT [16] discards too much context structure, resulting in cluttered images. Because it adjusts the overall second-order statistics of the content features to match style images, which will lead to ignoring content and style correlations (e.g., row 1 and row 6 in Fig.5). Avatar Net [18] utilizes multi-scale style transfer to transform content features into semantically closest style features while minimizing the difference between their overall features. Due to its dependency on the patch size, it hard to keep both local and global style patterns (e.g., row 2 and row 8 in Fig.5). SANet [14] uses Style-Attentional to match the style features to the content features to generate attractive results. Based on the SANet[14], our method not only can implement different style modes such as global color distribution and local style patterns but also maintain the structure of the content images (e.g., Fig.4 and Fig.5).

Compare with SANet [14], our method can flexibly to learn different types of the style features and enhance the structure of content images on the basis of maintaining the image content. Our method can provide high contrast between foreground and background. Due to minimizing the global structure Loss provides the strong global structural information, which is reflected in the depth of different objects in the content image. In particular, our method produces stronger visual effects in close-up images, such as human faces (e.g., Fig.4 and Fig.5). In addition, our method can capture the minute features by minimizing the local structure loss, which can more focus on the local information and refine the local details of the style, such as eyes detail (e.g.,row 1,4,5,8 in Fig.4). Complementary of the local and global loss, our method promotes the visual result. We also exhibit more details in the style pattern between different method in Fig.4 (such as the bush strokes of hair, the color distribution of style, the detail of the eyes and the texture of the face.)

In summary, the aim of our approach is balancing between the flexibility and quality for style transfer tasks. In the following section, we will also show that our method is also competitive in terms of quantitative evaluations.

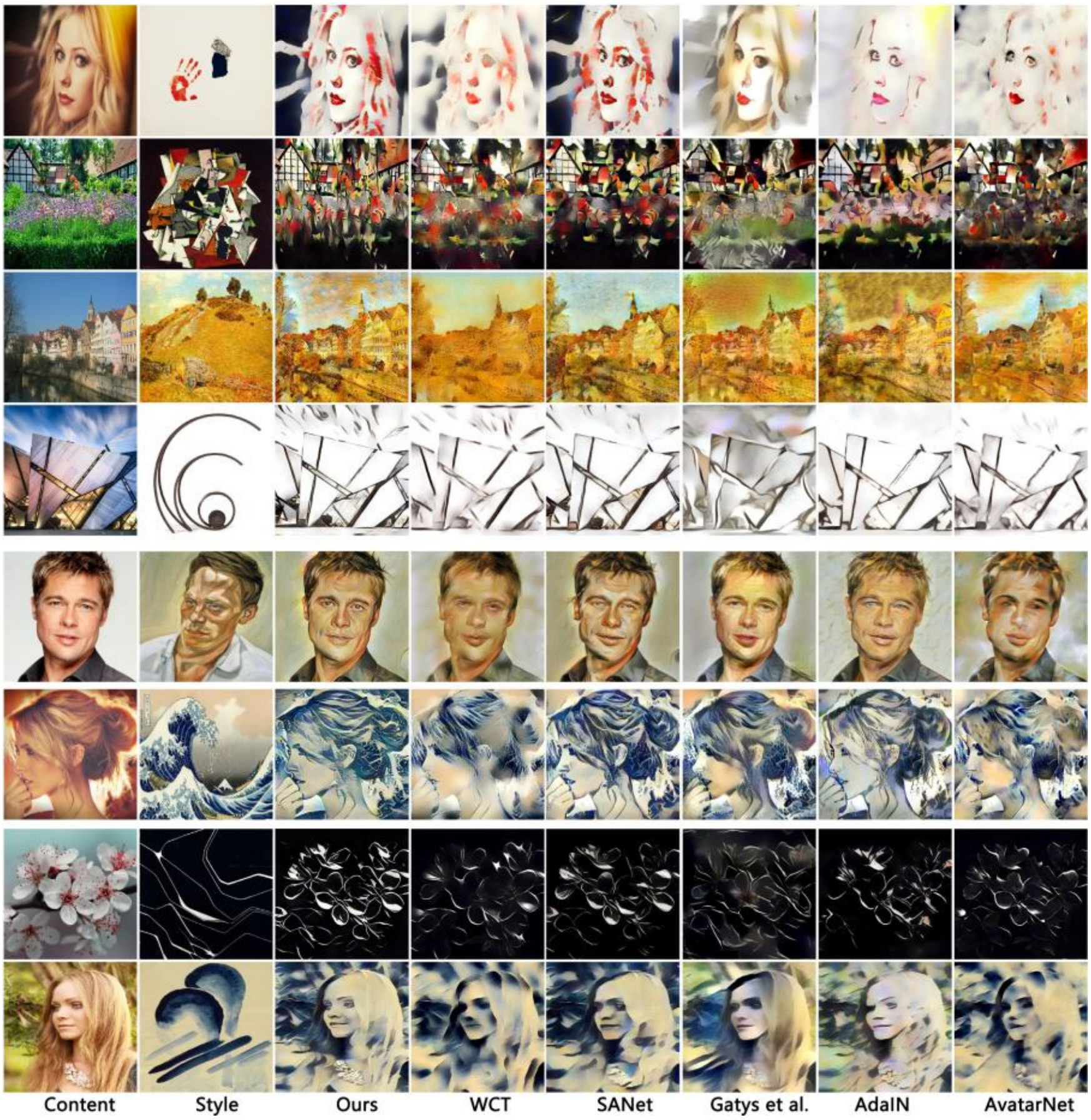

**Fig. 5** Example results for visual comparisons. Each row shares the same style, while each column represents the results of the same method.

#### 4.2.2 Quantitative Evaluations:

The evaluations of Style transfer is an open and challenging problem in this research. Due to the qualitative evaluations of style transfer is subjective, which is affected by evaluators' aesthetics and their experience. In order to make the evaluation more comprehensive, we do quantitative evaluations included the speed analysis ,structure consistency analysis and ablation analysis.

**Speed analysis.** As shown in Table.1,we compare our method with other arbitrary style transfer method [2][15][16][18] [14] on the run time performance respectively at two type images scales: $256 \times 256$ and $512 \times 512$ pixels. We measure the processing runtime of the testing including the style encoding procedure. Our method runs at 15 and 50 FPS for 256 x 256 and 512 x 512 image respectively, which speed make the possible to arbitrary user-selected styles in real times. Since our method is improved based on SANet, our approach has achieved the comparable speed to SANet [14] and AdaIN [15] (about 10% faster). Compared with [18] [16], our approach is 7-16 times faster than the patch-based method Avatar-Net [13] and 20-40 times faster than the feature transformation-based methods WCT [16]. Due to the limitation of iterative optimization-based method [2] itself, its computationally times is expensive. Therefore, our model is nearly 3 orders of magnitude faster than [2].

**Table 1** Execution time comparison (in seconds).

| Method | 256X256 | 512X512 | Styles |
|---|---|---|---|
| **Gatys et al.[2]** | 15.863 | 50.804 | ∞ |
| **WCT[16]** | 0.689 | 0.997 | ∞ |
| **Avatar-Net [13]** | 0.248 | 0.356 | ∞ |
| **AdaIN [15]** | 0.018 | 0.065 | ∞ |
| **SANet [14]** | 0.017 | 0.055 | ∞ |
| **Ours** | **0.015** | **0.050** | ∞ |

**Structure consistency analysis.**In addition to the visual quality, we also need to evaluate the structural consistency between the content images and the generated images. The depth map and edge map of the image can be used as an important basis for reflecting the image structure, which contains the depth information and edge information of the image. So we use the HED method [36] to extract the edge map of the content image and styled images, which generated by different style transfer methods, and extract the depth map of the corresponding image through the [29] method. As shown in Fig.6, our results maintain reasonable depth maps and Edge maps with structural and spatial consistency.

In addition, SSIM is an important indicator for measuring the structural similarity between two images. It evaluates the similarity of two images in terms of brightness, contrast and structure. So we use the SSIM [37] to evaluation the structural consistency. Noted that two images are nearly identical, their SSIM is close to 1. The Higher is better for SSIM (structural similarity).

**Table 2** Comparison on the structure consistency (SSIM)

| Method | Content SSIM | Depth Map SSIM | Edge Map SSIM |
|---|---|---|---|
| **Ours** | 0.410 | 0.886 | 0.477 |
| **SNAT [14]** | 0.364 | 0.846 | 0.423 |
| **AdaIN [15]** | 0.170 | 0.733 | 0.306 |
| **Avatar-Net [13]** | 0.215 | 0.763 | 0.420 |
| **WCT [16]** | 0.227 | 0.803 | 0.419 |
| **Gatys et al. [2]** | 0.195 | 0.683 | 0.319 |

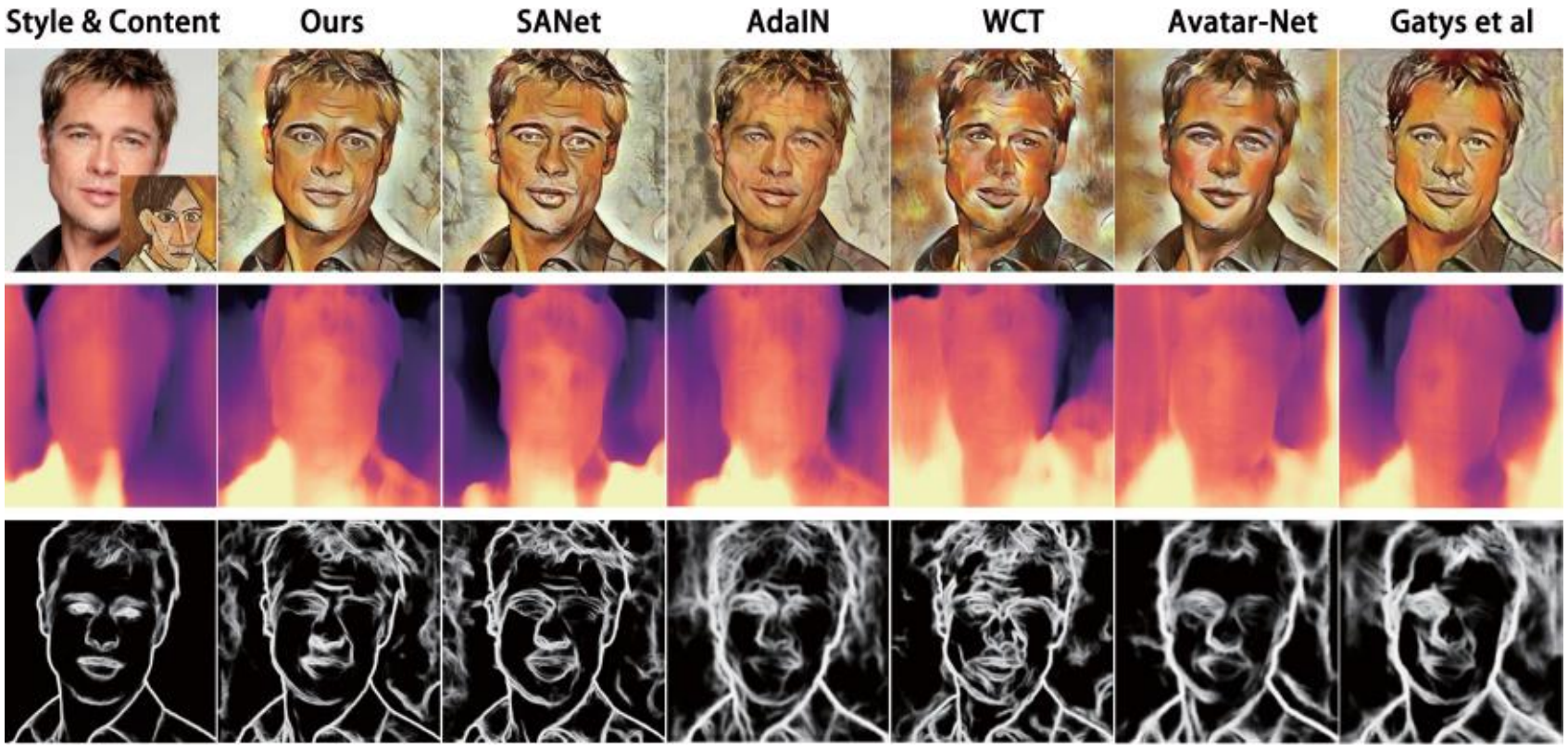

**Fig. 6** Comparisons on the structure consistency with other style transfer methods, the image depth map structure generated by our method is more prominent, and the edge structure preservation is more similar to the content image. The third row is the depth map using method [29]. The fourth row is the edge map using method [36].

Compare with other method, our proposed method has the highest SSIM value, whether it is the content image or the depth map and edge map. In particular, the SSIM value of my content map is qualitatively improved compared to other methods as shown in Table.2.

**Ablation Analysis.** In this section, we show the influence of the local loss and the global loss. As shown in Fig.7 (b) show the results obtained by fixing $\lambda_c$, $\lambda_s$, $\lambda_{Lap}$ at 1,3,0 while increasing $\lambda_{Depth}$ from 10 to 30 . As show in Fig.7(a) shows the results obtained by fixing $\lambda_c$, $\lambda_s$, $\lambda_{Depth}$ at 1, 3, 0, respectively, and increasing $\lambda_{lap}$ from 0.01 to 0.1 . When the style loss and content loss are fixed and if we increase the weight of the lap loss, the local details of the image will increase, but the overall visual feeling is still messy. Similarly, if we increase the weight of the deep loss, the global structure of the image will increase, but the local details will be need to improve. Therefore, we combine the depth loss and the Lap loss to enhance the content structure while enriching the style patterns (As shown in Fig.7.(c)).

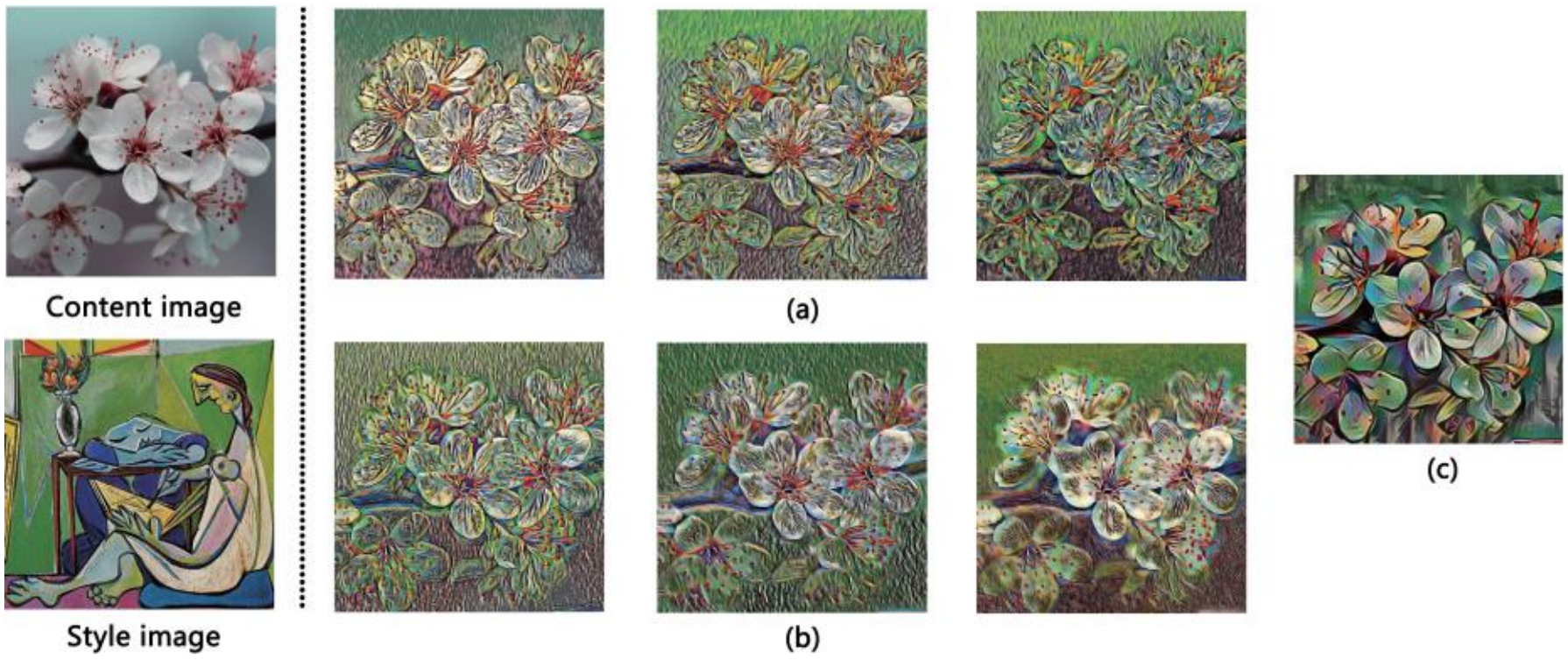

**Fig. 7** The first row (a) shows the results with only local loss.The second row (b) shows the results with only the global loss. The figure (c) shows the result with combining the global and local Loss.

## 4.3 Runtime Controls

In this section, we will show the flexibility application of our method, which allows the user to control the degree of stylization, the different Style interpolation, and the Spatial control. All these controls are only applied on the same network testing and not need to modify the training.

**Content–style trade-off.** During training, we can control the degree of stylization by adjusting the style weight $\lambda_s$ in Eq.5. In addition, we can control the degree of stylization by interpolating between feature maps that are fed to the decoder during testing. Note that this is equivalent to we adjust the stylized features, follow as:

$$\mathrm{T}(c, s, \alpha) = (1-\alpha)F_{cc} + \alpha F_{cs}^{''} \qquad \forall \alpha \in [0,1] \tag{12}$$

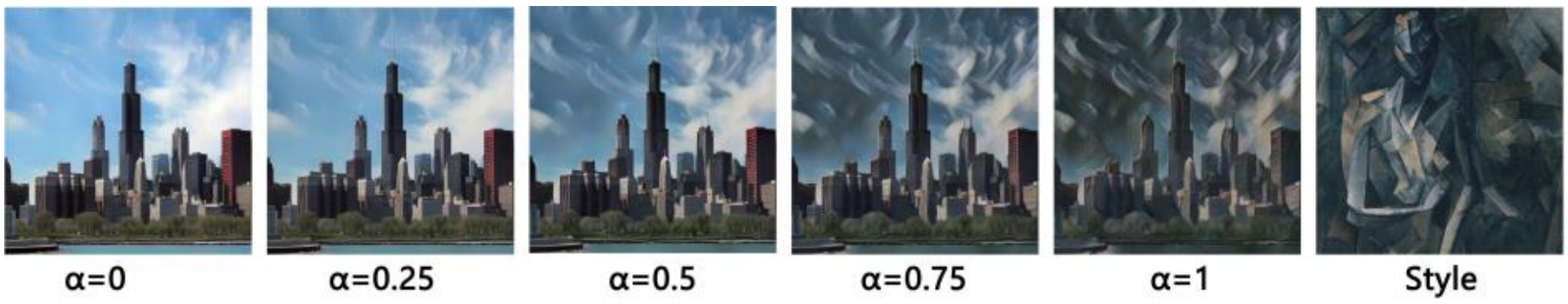

**Fig. 8** At runtime, we can control the balance between content and style by changing the weight $\alpha$ in Eq.12. The value of $\alpha$ is 0-1. When the $\alpha$ is larger, the degree of stylization is stronger.

Our method model takes two same content image as input to achieve Feature map $F_{cc}$. When $\alpha = 0$, the network tries to reconstruct the content image. When $\alpha = 1$, the network generates the most stylized image. As shown in Fig.8, we can achieve the style transition between content-similarity and style-similarity images when change the $\alpha$ from 0 to 1.

**Style interpolation** As shown in Fig.9, the different styles of feature maps $F_{cs}$ convex combing can be fed into the decoder to interpolate between different style images.

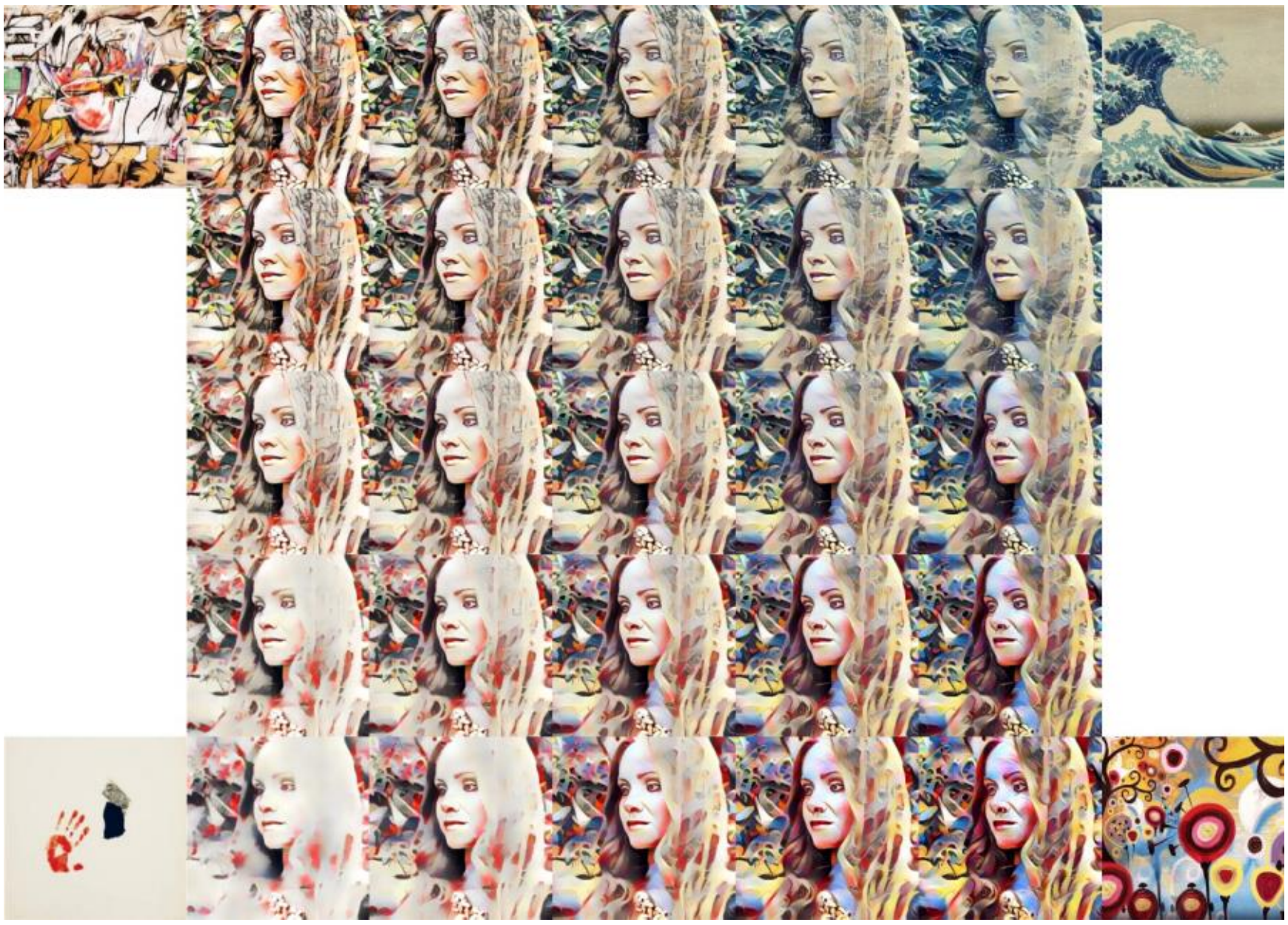

**Fig. 9** Example of style interpolation with four different styles.

**Spatial control** As shown in Fig.10, we demonstrate that our method can achieve the different regions of the content image to different stylization. The additionally input is a set of masks M (Fig.10 column 3), which maps the spatial correspondence between content regions and styles. The different styles can be assigned in each spatial region by replacing $F_{cs}''$ with M $\odot$ $F_{cs}''$, where $\odot$ is a simple mask out operation.

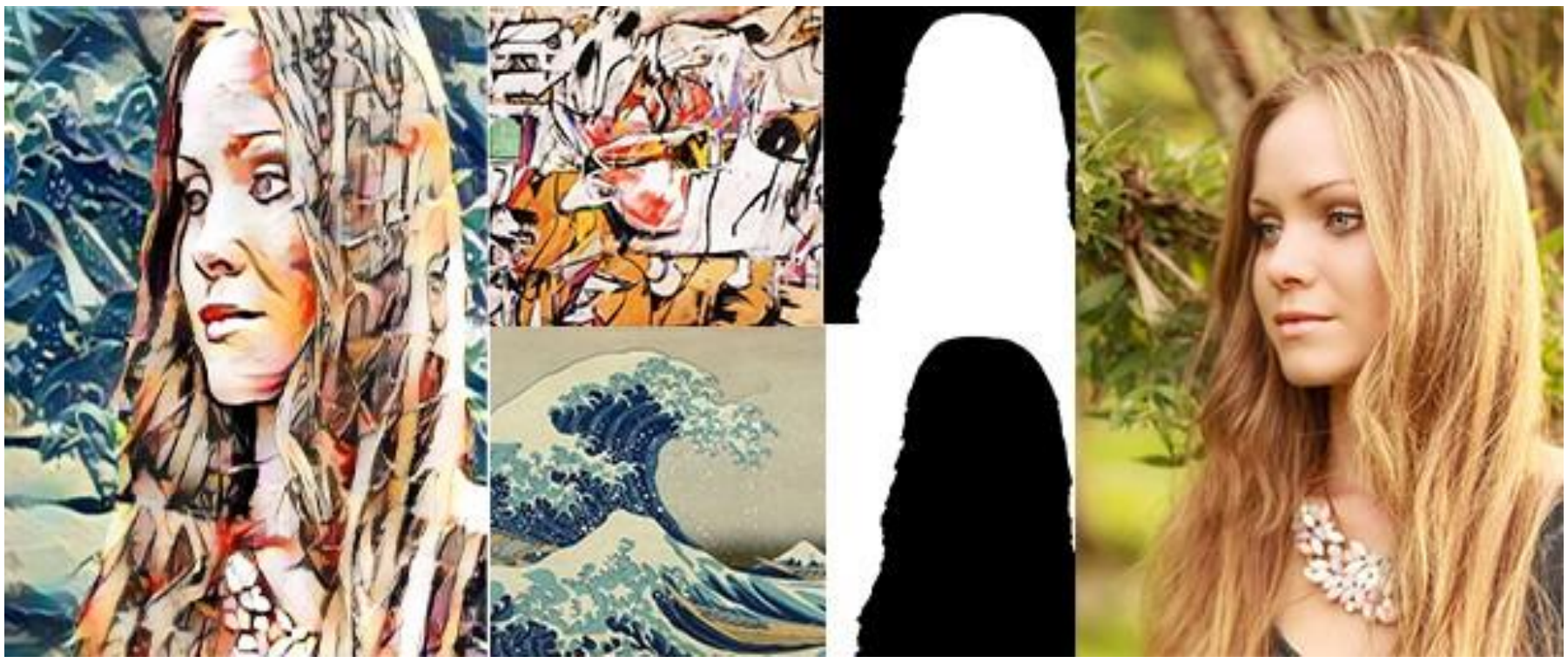

**Fig. 10** Example of spatial control. Left: Stylized image from two different style images. Middle: Style images and masks. Right: Content image.

# 5 Conclusion

In this work, we proposed a novel structure-enhanced image framework for arbitrary style transfer under the guidance of a combination of global structure loss and a local structure loss. We successfully applied arbitrary artistic effects in one trained model while retain the strong structure of images (specially in human face images). Experiment results demonstrate that our method is effective and efficient. Compared with other state-of-the-art methods, our method can maintain more content structure details and limited artifacts while still achieves an impressive visual result. Meanwhile, our method can be easily adapted in video arbitrary style transfer for the future work.

**Data availability.** Data sharing not applicable to this article as no datasets were generated during the current study.
**Conflict of interest.** The authors declare that they have no conflict of interest.